\documentclass[lettersize,journal]{IEEEtran}

\usepackage{amsmath,amsfonts}
\usepackage{algorithm}
\usepackage{algorithmic}

\usepackage{array}
\usepackage{multirow}
\usepackage{booktabs}
\usepackage{threeparttable}
\usepackage{adjustbox}
\usepackage[table]{xcolor}
\definecolor{gray15}{gray}{0.85}
\definecolor{gray51}{gray}{0.49}
\usepackage{pifont}

\usepackage{graphicx}
\usepackage[caption=false,font=footnotesize,labelfont=sf,textfont=sf]{subfig}

\usepackage{textcomp}
\usepackage{stfloats}
\usepackage{url}
\usepackage{verbatim}
\usepackage{float}
\usepackage{cite}

\usepackage{hyperref}

\usepackage{etoolbox}
\makeatletter
\patchcmd{\maketitle}{\@maketitle}{\vspace{-1.2em}\@maketitle}{}{}
\makeatother
\IEEEaftertitletext{\vspace{-0.8em}}

\title{
MM-DETR: An Efficient Multimodal Detection Transformer with Mamba-Driven Dual-Granularity Fusion and Frequency-Aware Modality Adapters
}

\author{Jianhong Han,~\IEEEmembership{Student Member,~IEEE},
        Yupei Wang$^{*}$,~\IEEEmembership{Member,~IEEE},
        Yuan Zhang$^{*}$,~\IEEEmembership{Member,~IEEE},
        and Liang Chen,~\IEEEmembership{Member,~IEEE}%
\thanks{This work was supported in part by the National Natural Science Foundation of China under Grant~62301046 and in part by the National Key Laboratory for Space-Born Intelligent Information Processing under Grant~TJ-01-22-01. (Corresponding authors: Yupei Wang and Yuan Zhang.)}%
\thanks{J. Han, Y. Wang, and L. Chen are with the School of Information and Electronics, Beijing Institute of Technology, Beijing 100081, China, and also with the Beijing Institute of Technology Chongqing Innovation Center, Chongqing 401135, China, and also with the National Key Laboratory for Space-Born Intelligent Information Processing, Beijing 100081, China (e-mail: hanjianhong1996@163.com; wangyupei2019@outlook.com; chenl@bit.edu.cn).}%
\thanks{Y. Zhang is with the School of Automation, Beijing Institute of Technology, Beijing 100081, China (e-mail: zhangyuan14@bit.edu.cn).}%
\thanks{Manuscript submitted to IEEE Transactions on Geoscience and Remote Sensing.}%
}

\begin{document}
\maketitle

\begin{abstract}
Multimodal remote sensing object detection aims to achieve more accurate and robust perception under challenging conditions by fusing complementary information from different modalities. However, existing approaches that rely on attention-based or deformable-convolution fusion blocks still struggle to balance performance and lightweight design. Beyond fusion complexity, extracting modality features with shared backbones yields suboptimal representations due to insufficient modality-specific modeling, whereas dual-stream architectures nearly double the parameter count, ultimately limiting practical deployment. To this end, we propose MM-DETR, a lightweight and efficient framework for multimodal object detection. Specifically, we propose a Mamba-based dual-granularity fusion encoder that first reformulates global interaction as channel-wise dynamic gating and leverages a 1D selective scan for efficient cross-modal modeling with linear complexity. Following this design, we further reformulate multimodal fusion as a modality-completion problem. A region-aware 2D selective-scanning completion branch is introduced to recover modality-specific cues. This branch further supports fine-grained fusion along a bidirectional pyramid pathway with minimal overhead. To further reduce parameter redundancy while retaining strong feature extraction capability, a lightweight frequency-aware modality adapter is introduced and inserted into the shared backbone. This adapter employs a spatial–frequency co-expert structure to capture modality-specific cues, while a pixel-wise router dynamically balances expert contributions for efficient spatial–frequency fusion. Extensive experiments conducted on four multimodal benchmark datasets demonstrate the effectiveness and generalization capability of the proposed method. Code is released at \href{https://github.com/h751410234/MM-DETR}{https://github.com/h751410234/MM-DETR}.

\end{abstract}


\vspace{2cm}

\section{Introduction}
\IEEEPARstart{D}{eep} learning–based methods for object detection have developed rapidly, aiming to achieve accurate classification and localization of multi-scale objects under complex backgrounds. However, the commonly used optical (RGB) modality suffers from significant performance degradation under adverse conditions such as low illumination, glare, smoke, or fog. In contrast, infrared (IR) imaging provides stronger robustness to these challenging conditions, but its lack of rich texture and color details makes it difficult to reliably detect small or subtle objects in remote sensing imagery. To mitigate these drawbacks, multimodal object detection has emerged as a research hotspot. By complementarily fusing features extracted from paired optical and infrared images, detectors can improve detection performance and achieve greater robustness across day–night scenarios and diverse weather conditions.

Although multimodal detection methods have made gradual progress, existing approaches still encounter bottlenecks in fusion mechanisms. To better exploit complementary information from the two modalities, recent studies\cite{yuan2024c2former,10669167,fu2024cf,liu2025dual} mainly rely on multi-head attention or deformable convolutional blocks to enable intensive interaction and alignment across multi-level features. While such designs facilitate cross-modal feature integration and improve detection performance, they also incur extremely high computational costs, which severely hinder model deployment and efficient inference in resource-constrained environments. Other works\cite{zhou2020improving,DING2021106990,xiao2024gm,liu2025aerial} have considered lightweight designs, primarily adopting FPN-style concatenation or convolution-based fusion. However, these approaches lack explicit alignment and selective complementary mechanisms, making them susceptible to information redundancy and feature misalignment, thereby limiting the full potential of multimodal complementarity.

Another fundamental challenge arises in feature extraction. To more effectively extract information from visible and infrared modalities, existing methods\cite{sharma2020yolors, yuan2024c2former , 10770223, zhu2025wavemamba, liu2025dual} often adopt dual-stream backbones to separately process the two inputs. Since backbone networks typically account for the majority of model parameters in lightweight detectors\cite{zhao2024detrs} (about 50\%), this dual-stream design nearly doubles the parameter scale and introduces additional bandwidth and latency overhead. As a result, it exacerbates the trade-off between accuracy and deployment, making it clearly unsuitable for lightweight detectors. An alternative approach\cite{zhang2023superyolo, xiao2024gm,liu2025aerial} is to share backbone parameters to reduce resource consumption and complexity. However, under highly heterogeneous modalities, shared representations are prone to conflicts, which may lead to representation collapse in one modality or mutual compromise across both, ultimately resulting in suboptimal performance. Thus, how to retain modality-specific cues under a shared backbone without falling back to heavy dual-stream architectures remains underexplored.

To address the aforementioned challenges, we propose MM-DETR, an effective and efficient detection transformer for multimodal object detection. Our approach integrates a Mamba-based Dual-granularity Fusion Encoder (MDF-Encoder) and inserts several proposed Lightweight Frequency-aware Modality Adapters (LFM-Adapters) into the backbone. Together, these components enhance and complement the effective fusion of multimodal features, while extracting modality-specific discriminative representations without relying on a dual-stream paradigm. Consequently, MM-DETR offers a favorable balance among accuracy, inference speed, and lightweight deployment.

To unify cross-modal commonality enhancement and complementary information fusion within a lightweight paradigm, we propose a novel \textbf{MDF-Encoder}. Specifically, this encoder introduces a Commonality-Enhancing Interaction (CEI) module, which reformulates cross-modal global interaction modeling as a channel-wise dynamic gating process. Built upon the One-Dimensional Selective Scan (SS1D) in Mamba\cite{gu2024mamba}, this module efficiently captures input-dependent global interactions with linear complexity, thereby enhancing shared representations across modalities. On this basis, we further develop the Modality-completion Pyramid Fusion (MPF) module, which redefines cross-modal feature fusion as a modality completion problem by integrating a Region-aware SS2D modality-completion branch. By dynamically compensating for modality-specific information along a bidirectional pyramid pathway\cite{liu2018path}, the MPF module achieves lightweight yet integrated multimodal feature fusion under complex remote sensing scenarios.

To enable efficient and discriminative multimodal representation learning without adopting a dual-backbone architecture, we propose a \textbf{LFM-Adapter}, which is integrated into each stage of the shared backbone. The adapter employs a spatial–frequency co-expert structure to jointly model modality discrepancies across spatial and frequency domains, thereby capturing modality-specific representations that reflect texture, structural, and energy distribution characteristics. Subsequently, a pixel-wise router dynamically balances the contributions of different experts in response to the input features, enabling adaptive coordination and fusion of spatial–frequency information. With this collaborative design, the proposed adapter enables efficient and adaptive multimodal feature extraction with minimal parameter overhead.

The main contributions of this paper are as follows:

\begin{itemize}
\item{We propose the MDF-Encoder that explicitly models cross-modal commonality enhancement and complementary information fusion in a lightweight fashion. It reformulates global feature modeling into a channel-wise dynamic gating process, achieving efficient cross-modal interaction with linear complexity. Additionally, a region-aware SS2D–based modality-completion branch is integrated into a bidirectional pyramid pathway, dynamically restoring missing or weak modality-specific cues and enabling lightweight yet highly effective multimodal feature fusion.}

\item{An LFM-Adapter is proposed that allows us to abandon the dual-stream backbones widely used in prior multimodal detectors, while still enabling efficient multimodal representation learning. The adapter integrates a spatial–frequency co-expert structure to capture modality-specific spatial and spectral characteristics, while a pixel-wise router adaptively balances expert contributions for dynamic coordination and efficient spatial–frequency fusion.}

\item{Extensive experiments conducted on multiple multimodal object detection benchmarks demonstrate the superior effectiveness and robustness of our MM-DETR, achieving 82.31\% mAP on DroneVehicle\cite{9759286} and 73.39\% mAP on M3FD\cite{liu2022target}, while maintaining a highly favorable accuracy–speed trade-off through a lightweight and efficient design.}

\end{itemize}

\section{Related Work}

\subsection{Multimodal Object Detection in Remote Sensing}

Multimodal remote sensing object detection aims to integrate complementary information from heterogeneous modalities such as optical and infrared imagery, thereby achieving more accurate and robust performance under challenging conditions such as low illumination or smoke. Given the procedural similarity between multimodal and unimodal detection, most mainstream methods adopt a dual-stream backbone architecture with independent parameters to extract modality-specific features, which are then integrated through carefully designed fusion modules to achieve cross-modal representation. Sharma et al.~\cite{sharma2020yolors} first performed multimodal fusion at mid-level feature stages within the YOLO framework, significantly improving detection accuracy. Zhang et al.~\cite{zhang2021guided} developed a cross-modal convolutional enhancement module based on the single-stage RetinaNet framework\cite{lin2017focal}. Liu et al.~\cite{liu2025dual} focused on cross-modal feature alignment, realizing multi-level feature consistency within the FCOS detector\cite{tian2019fcos}. In addition, classical two-stage frameworks such as Faster R-CNN~\cite{cao2023multimodal,yuan2024c2former} and Transformer-based detectors~\cite{guo2024damsdet,fu2024cf,helvig2024caff} have been extended to multimodal scenarios, continuously driving progress in cross-modal detection performance.

Although dual-stream architectures offer strong representational flexibility, their additional parameters hinder deployment on resource-constrained platforms. Recent studies have shifted toward shared-backbone architectures across modalities, achieving lightweight and efficient inference. Zhang et al.~\cite{zhang2023superyolo} realized fast multimodal detection via pixel-level concatenation fusion, whereas Liu et al.~\cite{liu2025aerial} proposed a multibranch progressive fusion mechanism based on lightweight convolutional modules. With the advent of real-time DETR-based detectors~\cite{zhao2024detrs}, Xiao et al.~\cite{xiao2024gm} extended the approach to multimodal object detection, achieving superior detection accuracy and efficiency, thereby highlighting the potential of Transformer architectures for lightweight multimodal detection. 

In this paper, we present an effective and efficient multimodal detection framework, termed MM-DETR, which enhances multimodal feature extraction and complementary information fusion, achieving a favorable balance between accuracy and efficiency.

\subsection{Feature Fusion in Multimodal Detection}

Feature fusion is a crucial design component in multimodal object detection, aiming to effectively integrate heterogeneous features to achieve more robust and discriminative representations. Owing to the popularity and effectiveness of attention mechanisms, most recent studies have adopted cross-modal attention modeling to establish fine-grained semantic consistency and spatial alignment between different modalities. Cao et al.~\cite{cao2023multimodal} proposed a spatial attention-based fusion strategy that adaptively aggregates complementary modal features. Zhang et al.~\cite{xu2024airborne} developed a multi-scale adaptive fusion module to enhance the discriminability and robustness of small objects. Zhu et al.~\cite{dong2024seadate} introduced a dual-attention mechanism that employs Transformer-based fusion blocks to jointly model spatial and channel dependencies. However, attention computation on multi-scale dense feature maps often imposes a heavy computational burden, making it difficult to balance global modeling capability and computational efficiency.

Another important paradigm for multimodal feature fusion leverages deformable convolution to achieve cross-modal alignment and information integration by adaptively adjusting convolutional sampling locations. Yuan et al.~\cite{yuan2024c2former} incorporated deformable convolution to realize both geometric and semantic alignment. Extending this idea, Zhou et al.~\cite{chen2024weakly} developed an adaptive feature alignment module based on deformable convolution to achieve robust spatial alignment in weakly registered imagery. Liu et al.~\cite{liu2025dual} introduced a hierarchical deformable aggregation mechanism within a feature pyramid network to accomplish dual-perspective alignment. Compared with attention-based approaches, deformable convolution-based fusion methods are more compact and efficient but still struggle to capture long-range dependencies and global semantic correlations across modalities.

Despite these advances, the trade-off between accuracy and efficiency remains a major challenge, prompting a growing interest in lightweight multimodal fusion. Early approaches typically relied on simple feature concatenation or element-wise operations to merge multimodal representations within the backbone~\cite{choi2016multispectral,sharma2020yolors} or the feature pyramid~\cite{zhou2020improving,sun2022drone}, aiming to reduce computational complexity. To further enhance cross-modal interaction, Xiao et al.~\cite{xiao2024gm} proposed GM-DETR, which performs cross-modal multi-scale fusion within a Transformer framework by leveraging its inherent self-attention mechanism. Liu et al.~\cite{liu2025aerial} employed stacked pooling and convolution operations to deepen feature interaction while maintaining a lightweight design. Nevertheless, these methods remain essentially local, struggling to capture long-range cross-modal dependencies and to fully exploit cross-modal complementarity. Inspired by the Mamba architecture, Zhu et al.~\cite{zhu2025wavemamba} employed state-space modeling to achieve global cross-modal interaction with lower computational complexity. Similarly, Shen et al.~\cite{SHEN2026103895} integrate multiple Mamba-based branches into the fusion neck to capture complementary and shared RGB–IR cues. Although these approaches effectively enhance multimodal interaction, they still rely on dense 2D parameterization over multi-scale high-resolution feature maps, which imposes substantial parameter overhead and significantly increases inference latency.

To achieve efficient cross-modal interaction and fusion with lower computational burden, we propose a Mamba-based Dual-granularity Fusion Encoder (MDF-Encoder). At the global level, we reformulate cross-modal interaction as a channel-wise dynamic selective scan, rather than operating on dense spatial tokens as in previous Mamba-based fusion blocks. At the regional level, we further reinterpret multimodal fusion as a modality-completion problem and introduce a Region-aware SS2D completion branch. By combining grouped low-rank state-space modeling with localized contextual cues, this branch dynamically restores missing or weak modality-specific information along a bidirectional pyramid pathway while maintaining minimal computational overhead.

\subsection{Adapter-Based Representation Learning}

Adapters have become a prominent paradigm of Parameter-Efficient Fine-Tuning (PEFT), widely adopted in natural language processing. They insert lightweight trainable modules into large-scale pretrained models, enabling task-specific adaptation with minimal additional parameters~\cite{houlsby2019parameter,hu2022lora}. For vision tasks, adapters have been further extended to convolutional neural networks (CNNs) and Vision Transformers (ViTs) to improve model transferability and feature representation~\cite{chen2022adaptformer,zhang2021tip,yin20255}.

Building upon these advances, several studies have further explored joint-tuning strategies that update both adapter modules and backbone parameters to achieve greater representational flexibility. Chen et al.~\cite{chen2022vision} jointly optimize adapter modules with backbone layers to enhance the adaptability of dense prediction models. Xie et al.~\cite{xie2023difffit} update a subset of backbone parameters alongside adapters, thereby achieving stronger transferability across diverse visual domains. Bhattacharjee et al.~\cite{bhattacharjee2023vision} further employ coordinated optimization across diverse multitask settings.

In parallel, researchers have incorporated Mixture of Experts (MoE) mechanisms into adapter design to further enhance model capacity for dynamic feature modeling and multi-task collaborative optimization. Wang et al.~\cite{wang2020deep} extend this concept to convolutional networks by treating convolutional channels as experts. Riquelme et al.~\cite{riquelme2021scaling} employ a gating mechanism to adaptively allocate information flow among multiple experts. Jain et al.~\cite{jain2024mixture} integrate MoE structures into Vision Transformers with image-level dynamic routing, enabling adaptive feature allocation and sharing across tasks. Yin et al.~\cite{yin20255} further introduce multiple vision-friendly expert filters within the adapter, strengthening its capability to process diverse visual signals.

Inspired by the aforementioned approaches, we propose the Lightweight Frequency-Aware Modality Adapter (LFM-Adapter) to efficiently extract features from different modalities with minimal additional parameters. This adapter integrates a spatial–frequency co-expert module with a pixel-wise router, enabling finer modality-specific feature representation learning and adaptive modality fusion while maintaining high parameter efficiency. To the best of our knowledge, LFM-Adapter is among the earliest attempts that explicitly explore the trade-off between backbone parameter efficiency and detection performance in multimodal object detection.

\section{Method}

\subsection{Preliminaries}

\subsubsection{Selective Scan in One Dimension}

To efficiently extend the continuous-time state-space formulation into a discrete and learnable mechanism, Mamba introduces the Selective Scan in One Dimension (SS1D). This module performs a causal scan along the sequential dimension, dynamically updating the latent state through input-dependent transition matrices. Compared with the fixed-parameter linear recurrence in standard SSMs, SS1D adaptively modulates the information flow at each step, thereby capturing both long-term dependencies and input-conditioned dynamics with linear-time efficiency.

Given an input sequence $\mathbf{U} = \{\mathbf{u}_1, \mathbf{u}_2, \dots, \mathbf{u}_L\}$, the hidden state $\mathbf{h}_t$ and output $\mathbf{y}_t$ are updated as:
\begin{equation}
\mathbf{h}_t = \bar{A}_t\, \mathbf{h}_{t-1} + \bar{B}_t\, \mathbf{u}_t, 
\quad
\mathbf{y}_t = C_t\, \mathbf{h}_t,
\end{equation}
where $\bar{A}_t$, $\bar{B}_t$, and $C_t$ denote the input-dependent coefficients obtained through lightweight linear projections. Such parameterization allows the model to selectively integrate or forget contextual information depending on the semantic importance of each token.

Unlike self-attention, which explicitly computes all pairwise dependencies 
with a complexity of $\mathcal{O}(L^2)$, SS1D achieves implicit 
long-range modeling through a recursive selective update, maintaining 
both $\mathcal{O}(L)$ computational and memory efficiency. 
Building upon the original Mamba scan, which performs recurrence along the sequential dimension, we reformulate SS1D as a feature-dependent gating mechanism that adaptively regulates information flow across modality channels, thereby providing a lightweight yet effective paradigm for cross-modal interaction modeling.

\subsubsection{Selective Scan in Two Dimensions}
\label{sec:ss2d}

To extend selective state-space modeling from one-dimensional sequences to the image domain for visual tasks, 
the Selective Scan in Two Dimensions (SS2D) treats an input feature map as an ordered set of sequences. 
Specifically, the feature map is unfolded along a specific scan direction to preserve the spatial continuity of image structures. 
Each scanning path independently performs an SS1D operation, and the responses from all paths are subsequently folded back into a two-dimensional layout, 
forming a structurally continuous and semantically coherent 2D representation. 
Let $\mathcal{D}$ denote the set of scan directions. 
The final 2D response is obtained by folding each directional output ${\mathbf{y}^{(d)}}$ back to $H\times W$ and aggregating them across directions as follows:
\begin{equation}
\mathrm{SS2D}(\mathbf{X})
=\mathrm{Fold}\!\left(\sum_{d\in\mathcal{D}}\mathrm{SS1D}\!\big(\mathrm{Unfold}_d(\mathbf{X})\big)\right).
\end{equation}

\subsubsection{Region-aware SS2D} While SS2D generalizes SS1D to image-like feature maps through directional unfolding, its parameterization still relies on dense full-rank linear 2D mappings to generate state-transition matrices. This design becomes suboptimal for detection backbones because multi-scale pyramids force the same module to operate on feature maps of varying resolutions, making the full-domain $H\times W$ mappings incur a substantial constant factor in both parameters and FLOPs. Moreover, RGB–IR modality discrepancies exhibit strong spatial heterogeneity, making a region-adaptive state-space formulation particularly important in multimodal settings. To address this issue, we introduce Region-aware SS2D, a lightweight variant of SS2D that augments the original design with region-adaptive conditioning and grouped low-rank parameterization. Unlike existing Mamba-based vision models~\cite{liu2024vmamba,zhu2025wavemamba,SHEN2026103895} that rely on globally dense projections across spatial scales, our formulation explicitly enables region-level state modeling while preserving linear computational complexity. This design enables the module to more effectively handle multi-scale feature inputs in detection tasks and to further reduce the overall parameter overhead.

Given an input feature map $\mathbf{X}\in\mathbb{R}^{B\times C\times H\times W}$, Region-aware SS2D first extracts localized contextual cues through a lightweight depthwise convolution:
\begin{equation}
\tilde{\mathbf{X}} = \mathrm{SiLU}\big(\mathrm{DWConv}(\mathrm{GN}(\mathbf{X}))\big),
\end{equation}
where $\mathrm{GN}(\cdot)$ denotes Group Normalization and $\mathrm{DWConv}(\cdot)$ is a depthwise convolution that supplies region-aware local context. This region-wise normalization mitigates the scale-dependent statistical discrepancies across pyramid levels, allowing a single Region-aware SS2D block to operate robustly over multi-scale inputs even under multi-scale training without requiring any reparameterization.

After flattening the spatial dimensions into a sequence, Region-aware SS2D follows the general SS2D process to generate the driving signal $\mathbf{U}$ and step-size modulation $\Delta$. Unlike standard SS2D, which uses one dense projection of size $C\times C$, we adopt a pair of low-rank linear mappings that factorize the transformation into a compact bottleneck. This reduces complexity from $\mathcal{O}(C^{2})$ to $\mathcal{O}(Cr)$ with rank $r$, while enabling richer region-specific dynamics than a single low-rank layer. Empirically, we set $r = 4$ across all experiments, which provides a stable accuracy–efficiency trade-off. To further improve modeling flexibility under a fixed parameter budget, we adopt a grouped low-rank formulation for the state-transition and output matrices $\mathbf{B}$ and $\mathbf{C}$. The channel dimension is partitioned into two independent groups, each processed separately. Within each group, a group-wise $1\times 1$ convolution yields a low-rank bottleneck representation, which is then expanded by a second projection into the full set of state-space parameters. After generating all state parameters, Region-aware SS2D performs selective scans along the horizontal and vertical directions and folds the resulting sequences back into the two-dimensional spatial layout.

\begin{figure*}[t]
\centering
\includegraphics[width=18cm]{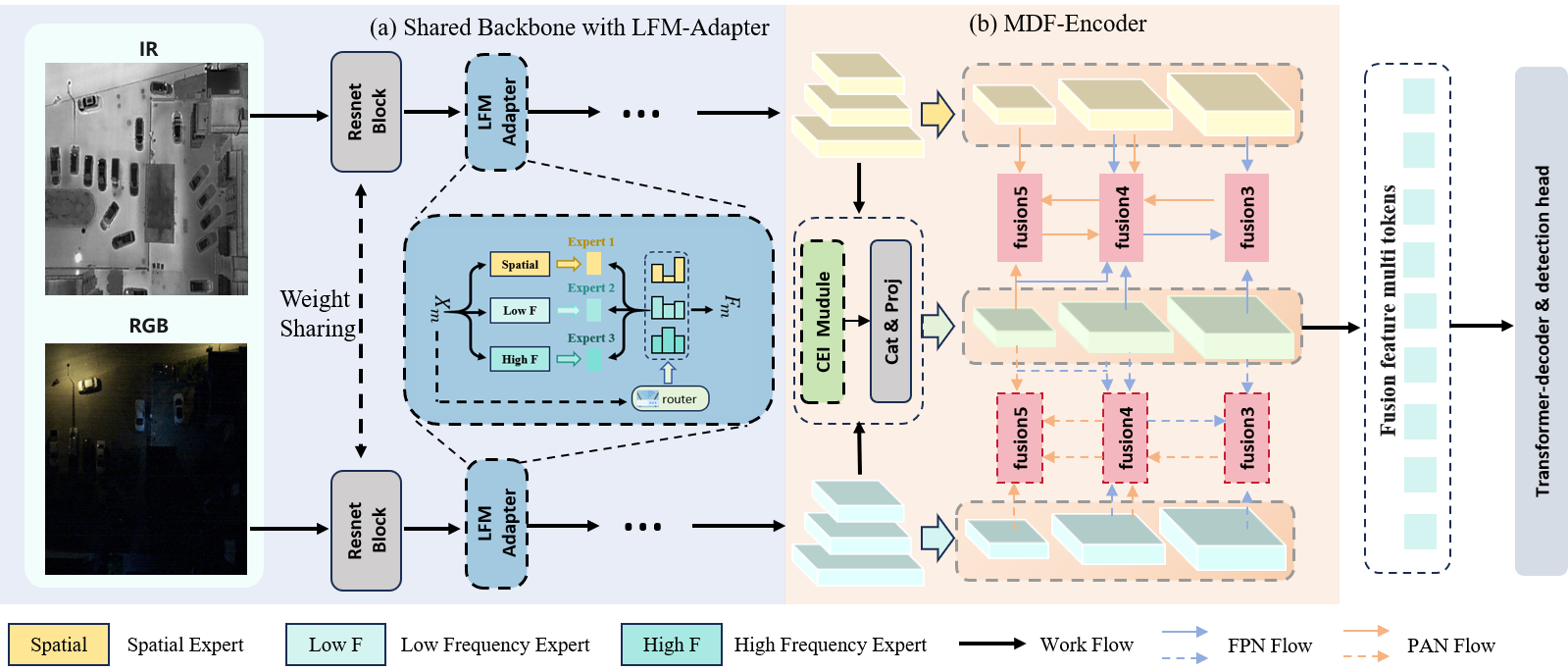}
\caption{Overall architecture of the proposed MM-DETR. Given paired RGB and IR images, a shared RT-DETR backbone equipped with lightweight frequency-aware modality adapters extracts modality-specific multi-scale features for both streams. These features are then fed into the Mamba-based Dual-granularity Fusion Encoder (MDF-Encoder). The MDF-Encoder first applies the commonality-enhancing interaction module to perform global cross-modal interaction, producing enhanced modality-shared representations. Subsequently, the modality-completion pyramid fusion module performs complementary fusion along a bidirectional pyramid, where the dashed arrows indicate an optional RGB-side completion path. Finally, the fused multi-scale tokens are passed to the transformer-decoder and detection head to generate the final predictions.}
\label{fig_1}
\end{figure*}

\subsection{Framework Overview}

As illustrated in Fig.~\ref{fig_1}, the proposed MM-DETR adopts the lightweight RT-DETR~\cite{zhao2024detrs} as its base detector, enabling efficient and effective multimodal object detection. For modality-specific feature extraction, we employ a shared backbone augmented with a LFM-Adapter (Sec.~\ref{Adapter}). The Adapter is dedicated to recovering modality-specific cues within the shared backbone, ensuring that RGB and IR features preserve their individual characteristics without relying on a dual-stream architecture. For cross-modal interaction and fusion, we introduce a MDF-Encoder (Sec.~\ref{Encoder}), which contains two functionally distinct modules. The commonality-enhancing interaction module operates at the global granularity, strengthening modality-shared semantics through SS1D-based dynamic gating. Then, the modality-completion pyramid fusion  module operates at the region granularity, injecting modality-specific residual information via region-aware SS2D along the bidirectional pyramid pathway.

Specifically, given a pair of visible and infrared images, we first employ a shared backbone to extract multi-scale features for each modality. For each stage $l$, the backbone produces feature maps denoted as $X_m^{l}$, where $m \in {\text{rgb}, \text{ir}}$ indicates the modality type. Subsequently, $X_m^{l}$ is fed into the corresponding modality-specific adapter to further refine the representation and capture modality-specific characteristics, yielding the output feature $F_m^{l}$. The resulting $F_m^{l}$ is then used as the input to the next stage, where the backbone and adapter continue the feature extraction process.
Finally, each modality produces three groups of multi-scale feature maps, denoted as $\{{F_{\text{rgb}}^{3}, F_{\text{rgb}}^{4}, F_{\text{rgb}}^{5}}\}$ and $\{{F_{\text{ir}}^{3}, F_{\text{ir}}^{4}, F_{\text{ir}}^{5}}\}$, respectively. These multimodal multi-scale feature maps are then fed into the proposed MDF-Encoder to perform efficient feature alignment and fusion. The decoder, prediction head, and objective function of MM-DETR follow the vanilla RT-DETR. It is worth noting that the training labels are defined as the union of the annotations from both modalities to ensure consistency with the multimodal image inputs.

\subsection{Mamba-based Dual-Granularity Fusion Encoder}
\label{Encoder}

\begin{figure}[t]
\centering
\includegraphics[width=9cm]{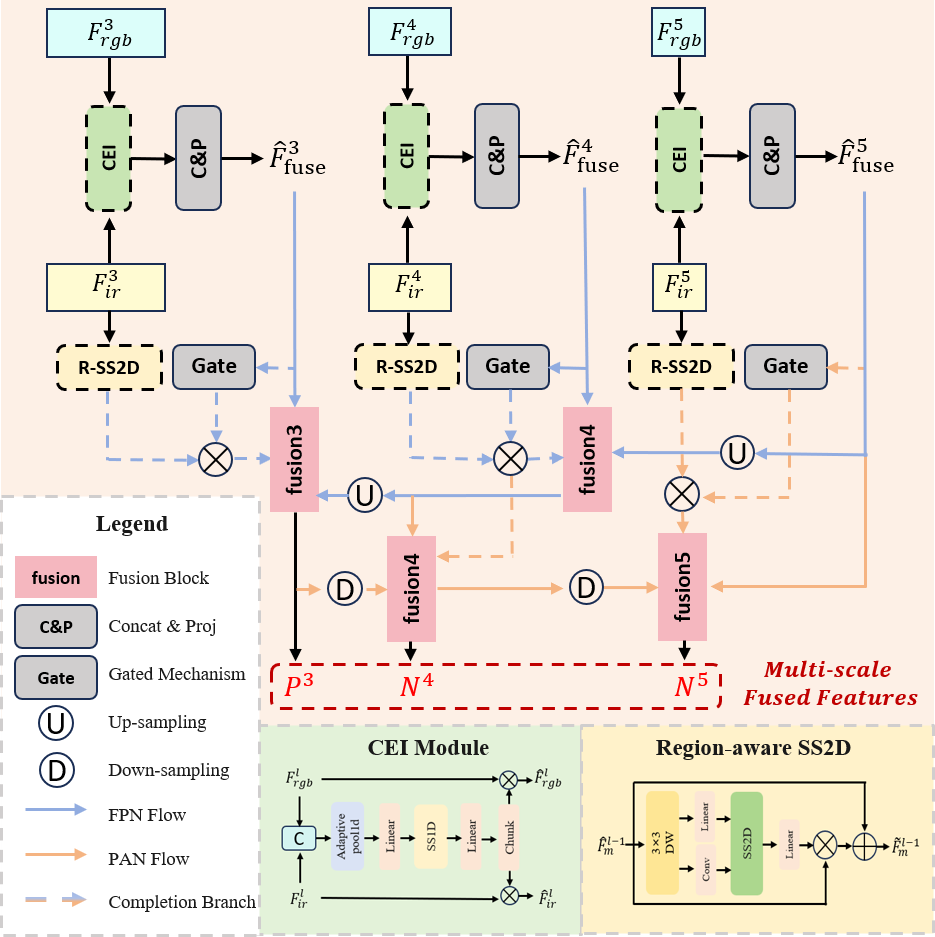}
\caption{Detailed architecture of the proposed Mamba-based Dual-Granularity Fusion Encoder (MDF-Encoder). At each pyramid level, the Commonality-Enhancing Interaction (CEI) module first performs channel-wise cross-modal interaction to reinforce modality-shared representations, producing the fused features. As illustrated by the dashed paths, an additional modality-completion branch is incorporated, where the Region-aware SS2D module enhances modality-specific residual cues from the IR stream and processes them through a lightweight gated mechanism to generate completion features. These completion cues are subsequently injected into both the top–down (FPN) and bottom–up (PAN) pathways via dedicated fusion blocks, enabling fine-grained complementary fusion across scales.}
\label{fig_2}
\end{figure}

Balancing computational and parameter efficiency against fusion effectiveness is crucial for the practical deployment of multimodal object detectors. Existing methods often mitigate modality misalignment and cross-modal interference by designing sophisticated fusion blocks that rely on multi-head attention or deformable convolutions, which inevitably introduce substantial computational overhead. In contrast, lightweight designs typically fuse features through depthwise-separable convolutions combined with concatenation-based operations, yet they still struggle under large modality discrepancies and fail to fully exploit cross-modal complementarity. Inspired by Mamba-based vision networks, we propose a Mamba-based dual-granularity fusion encoder, comprising the Commonality-Enhancing Interaction (CEI) and Modality-completion Pyramid Fusion (MPF) modules. The CEI employs SS1D-based channel-wise dynamic gating to strengthen cross-modal commonality at the global granularity, while the MPF performs modality-specific completion through region-aware SS2D along a bidirectional pyramid pathway at the regional granularity. Together, these two modules achieve explicit commonality enhancement and complementary fusion while maintaining a lightweight overall design.

\subsubsection{Commonality-Enhancing Interaction module} Unlike simply replacing attention with a generic Mamba block on spatial tokens,
which still suffers from complexity and memory usage tied to long spatial sequences,
our CEI module applies SS1D to a pooled channel sequence obtained from concatenated RGB–IR features and thus reformulates cross-modal long-range interaction as a channel-wise gating problem. By operating in this channel space, CEI explicitly enhances cross-modal commonality and suppresses redundant channels,
while avoiding the misalignment introduced by irregular spatial sampling. Specifically, let the features from the two modalities at the $l$-th stage be denoted as $F_{\mathrm{rgb}}^{l}$ and $F_{\mathrm{ir}}^{l} \in \mathbb{R}^{B \times C \times H_l \times W_l}$. We first concatenate them along the channel dimension and flatten the result to obtain the merged feature representation. To alleviate spatial-resolution inconsistencies across multi-scale features, we further apply adaptive average pooling to transform the merged feature into a fixed-length sequence, thereby mapping cross-scale spatial variations into a unified sequential coordinate space. The process can be formulated as:
\begin{equation}
\mathbf{Z}^{l} = \mathrm{AdaptivePool}\!\left(
    \mathrm{Cat}\!\left[
        F_{\mathrm{rgb}}^{l},\, F_{\mathrm{ir}}^{l}
    \right]
\right).
\end{equation}

Subsequently, we perform one-dimensional selective scanning on the concatenated sequence $\mathbf{Z}^{l}$. In contrast to conventional designs, we parameterize the expanded matrix using low-rank factorization, which effectively reduces the number of parameters while preserving the representation capacity:
\begin{equation}
\mathbf{U}^{l} = \mathrm{SiLU}\!\left(
    \mathrm{Linear}\!\left(
        \mathbf{Z}^{l}
    \right)
\right),
\end{equation}
where Linear (·) denotes a fully connected layer used for low-rank feature projection, and SiLU refers to the Sigmoid Linear Unit activation function. Following the Mamba formulation, the sequence feature is split into the driving and gating branches. The driving branch produces input-dependent step sizes and state parameters via a lightweight linear projection, after which a linear state-space recursion is performed along the sequence dimension to obtain $\mathbf{Y}^{l}$:
\begin{equation}
\mathbf{Y}^{l} = \mathrm{SS1D}\!\left(
    \mathbf{U}^{l}
\right),
\end{equation}
The  $\mathrm{SS1D}$ operation effectively mixes and captures channel-related features by scanning the channel-mapping representations of the two modalities. Furthermore, $\mathbf{Y}^{l}$ is normalized and linearly projected to obtain channel-wise weights for both modalities, which reflect the relative importance of each channel in the two modalities:
\begin{equation}
\big( \mathbf{W}_{\mathrm{rgb}}^{l}, \; \mathbf{W}_{\mathrm{ir}}^{l} \big) 
= \mathrm{split}\!\left[
    \sigma\!\left(
        \mathrm{Linear}\!\left(
            \mathrm{LN}\!\left(\mathbf{Y}^{l}\right)
        \right)
    \right)
\right],
\end{equation}
where $\mathrm{LN}(\cdot)$ denotes the layer normalization layer, and $\sigma(\cdot)$ denotes the sigmoid activation function. Finally, the resulting weights are injected back into each modality in a residual manner, enabling explicit alignment and complementary enhancement:
\begin{equation}
\left\{
\begin{aligned}
\hat{F}_{\mathrm{rgb}}^{l} 
&= F_{\mathrm{rgb}}^{l} 
+ W_{\mathrm{rgb}}^{l} \odot F_{\mathrm{rgb}}^{l}, \\[4pt]
\hat{F}_{\mathrm{ir}}^{l} 
&= F_{\mathrm{ir}}^{l} 
+ W_{\mathrm{ir}}^{l} \odot F_{\mathrm{ir}}^{l} .
\end{aligned}
\right.
\end{equation}

\subsubsection{Modality-Completion Pyramid Fusion module} The CEI module enhances modality-shared representations at the global semantic level, but cross-modal fusion also requires recovering modality-specific cues that may be degraded or missing in one modality. To address this, we introduce the MPF module, which focuses on region-level complementary fusion. Prior lightweight fusion approaches generally rely on static operations such as concatenation or depthwise convolutions, often causing modality imbalance when one modality dominates the other. Conversely, pixel-level interaction can provide more precise fusion but brings substantial computational overhead.

To achieve both effectiveness and efficiency, MPF reformulates cross-modal fusion as a modality-specific feature completion problem. Instead of densely blending two modalities, MPF injects carefully selected residual cues from the more reliable modality to compensate for modality discrepancies and restore fine-grained details along the bidirectional pyramid pathway. This design maintains strong complementarity while preserving a lightweight computational profile.

Although the MPF structure is conceptually symmetric for RGB and IR modalities (Fig.~\ref{fig_1} (b)), our empirical analysis in Sec.~\ref{sec:ablation_mpf_branch} shows that instantiating only the infrared modality-completion branch achieves a significantly better balance between parameter cost and detection accuracy. IR contributes more stable structural cues under adverse illumination, whereas RGB completion introduces redundant or conflicting information. Therefore, in the following description, we focus on the IR-side completion branch, as illustrated in Fig.~\ref{fig_2}.

Given $\{ \hat{F}_{\mathrm{rgb}}^{l}, \hat{F}_{\mathrm{ir}}^{l} \}$, 
where $l \in \{3, 4, 5\}$ denotes the multi-scale feature maps enhanced by the CEI module, 
we first rapidly construct the fused feature $\hat{F}_{\mathrm{fuse}}^{l}$. 
This is achieved by concatenating the two modality features along the channel dimension 
and applying a $1 \times 1$ convolution for dimensionality reduction and alignment, 
which can be formally expressed as:
\begin{equation}
\hat{F}_{\mathrm{fuse}}^{l} = \mathrm{Proj}\big(\mathrm{Cat}[\, \hat{F}_{\mathrm{rgb}}^{l},  \hat{F}_{\mathrm{ir}}^{l}\,]\big),
\quad l \in \{3, 4, 5\},
\end{equation}
where $\mathrm{Proj}(\cdot)$ represents a $1\times1$ convolution used for reduction and alignment of dimensions. After obtaining the fused features, the deepest representation isprocessed by a self-attention block to enhance global dependency modeling, thereby providing stable semantic support for the subsequent pyramid pathways.

Building on this foundation, the MPF module follows the PAN-like bidirectional pyramid design of RT-DETR and further incorporates an additional modality-completion branch. In the top-down pathway, the highest-level fused feature $\hat{F}_{\mathrm{fuse}}^{5}$ serves as the starting point, and semantic information is progressively propagated to lower layers while injecting modality-completion cues. Specifically, modality-specific features are first modeled through the Region-aware SS2D module to efficiently capture both local and global contexts, thereby enhancing modality representations. Subsequently, we design a lightweight gated mechanism based on depthwise separable convolution and a pooling layer, as illustrated below:
\begin{equation}
\left\{
\begin{aligned}
\tilde{F}_{\mathrm{ir}}^{\,l-1} &= \mathrm{R\text{-}SS2D}\!\left(\hat{F}_{\mathrm{ir}}^{\,l-1}\right), \\[4pt]
R_{\mathrm{td}}^{\,l-1} &= \sigma\!\big(\mathrm{GAP}(\hat{F}_{\mathrm{fuse}}^{\,l-1})\big)
\odot \mathrm{DWConv}\!\left(\tilde{F}_{\mathrm{ir}}^{\,l-1}\right),
\end{aligned}
\right.
\end{equation}
where R-SS2D(·) denotes the region-aware selective scan in two dimensions, GAP(·) represents the global average pooling, and 
DWConv(·) indicates a depthwise convolution. 
This design equips the modality-completion branch with the ability to dynamically regulate information flow 
so that reliable infrared residual signals are amplified and incorporated into the shared representation, 
whereas noisy responses are suppressed by the gating mechanism.

Finally, the higher-level fused feature $\hat{F}_{\mathrm{fuse}}^{\,l}$ is first upsampled and then concatenated with the fused backbone feature of the current layer $\hat{F}_{\mathrm{fuse}}^{\,l-1}$ and the modality-specific feature from the modality-completion branch $R_{\mathrm{td}}^{\,l-1}$. The concatenated features are subsequently fed into the fusion block for joint integration, as illustrated below:
\begin{equation}
P^{\,l-1} = \mathrm{A}^{\,l-1}\!\Big(
\mathrm{Cat}\!\big[
\mathrm{U}_{\,l}(\hat{F}_{\text{fuse}}^{\,l}),\;
\hat{F}_{\text{fuse}}^{\,l-1},\;
R_{\text{td}}^{\,l-1}
\big]\Big), \quad l \in \{5,4\},
\label{eq:topdown_fusion}
\end{equation}
where $\mathrm{U}^{l}(\cdot)$ denotes the upsampling operation, $\mathrm{A}^{l-1}(\cdot)$ represents the fusion block composed of a $1\times1$ convolution followed by several RepBlocks. This fusion process explicitly incorporates the modality-specific completion mechanism during the top-down semantic propagation, enabling the model to dynamically leverage the strengths of individual modalities and effectively mitigate the degradation of modality-specific representations under complex scenes.

After completing the top-down semantic propagation, the MPF module further performs a bottom-up path that feeds fine-grained structural details back into higher-level representations, thereby establishing bidirectional complementarity between semantics and details. In this process, a symmetric architectural design, mirroring the top-down pathway, is employed at each layer to explicitly introduce a modality-completion branch. As illustrated in Fig.~\ref{fig_2}, starting from $P_3$, we progressively generate $\{N_4, N_5\}$, during which the high-level features explicitly acquire spatial details from the lower-layer modality-specific representations through semantic feedback. Ultimately, the module outputs the multi-scale fused features $\{P_3, N_4, N_5\}$, which are subsequently fed into the Transformer-based decoder head to produce the final detection results.

\subsection{Lightweight Frequency-Aware Modality Adapter} 
\label{Adapter}

\begin{figure}[t]
\centering
\includegraphics[width=9cm]{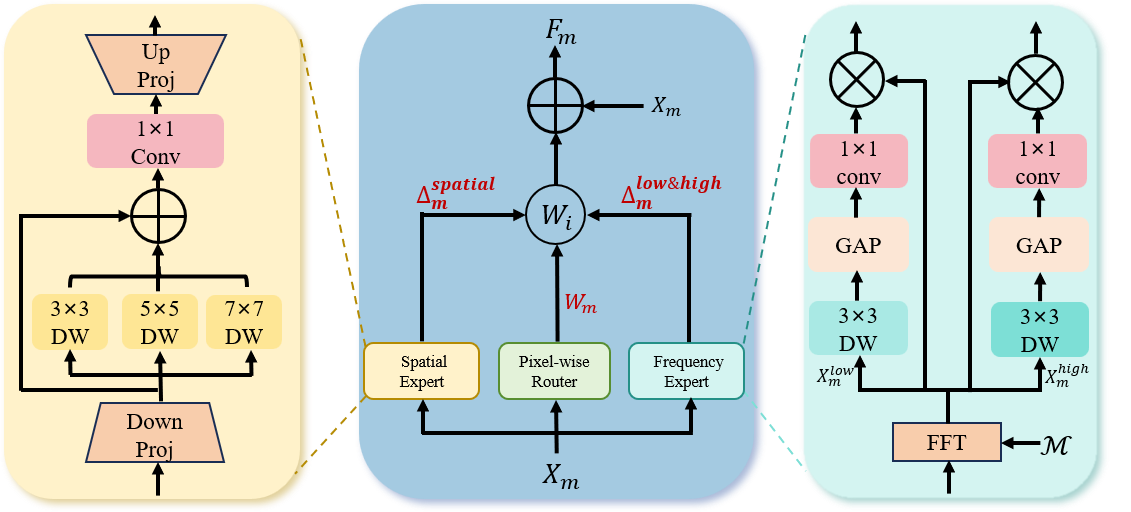}
\caption{Illustration of the proposed Lightweight Frequency-Aware Modality Adapter (LFM-Adapter). Left: the spatial expert structure. Right: the frequency expert structure. Middle: a pixel-wise router predicts adaptive weights to fuse the outputs of the spatial, low-frequency, and high-frequency experts.}
\label{fig_3}
\end{figure}

Due to the discrepancies between different modalities, existing methods typically adopt dual-stream backbones with identical architectures but unshared parameters to extract modality-specific features separately. Although such a design effectively captures modality-specific information, the resulting large number of parameters severely limits model deployment efficiency. Inspired by Parameter-Efficient Fine-Tuning (PEFT) techniques, which adapt large foundation models to various downstream tasks through the insertion of lightweight learnable modules, we further explore integrating a Mixture of Adapters (MoA) into a shared-backbone to enable efficient multi-modal feature extraction with minimal additional parameters. 

However, existing PEFT-based methods (e.g., Mona\cite{yin20255}) are primarily designed for natural-scene tasks and struggle to adequately model inter-modal discrepancies and complementarity, which leads to sub-optimal performance in multimodal detection. Considering that modality discrepancies often manifest as local variations in texture, brightness, and edge information, which are further amplified in the frequency domain as distinct energy distributions and structural patterns, we propose a LFM-Adapter. Specifically, the proposed LFM-Adapter adopts a collaborative design that integrates a spatial–frequency co-expert module with a pixel-wise router. The co-Expert module separately extracts modality-specific representations from both the spatial and frequency domains, while the router adaptively regulates the contribution of each expert during fusion according to the original visual features. To the best of our knowledge, LFM-Adapter is the first adapter framework that explicitly trades backbone parameter efficiency against modality-specific expressiveness by jointly leveraging spatial and frequency-domain experts under a shared-backbone multimodal detector.

\subsubsection{Spatial–Frequency Co-Experts} Let the input feature of modality $m \in \{\text{rgb}, \text{ir}\}$ at the $l$-th stage of the backbone be denoted as $X_m \in \mathbb{R}^{B \times C \times H_l \times W_l}$.
To enhance the numerical stability of multi-scale feature training, the input feature is first normalized in the spatial domain and then passed through a $1 \times 1$ convolution for a low-rank projection to refine the representation:
\begin{equation}
\tilde{X}_{m} = \mathrm{Conv}_{1\times1}\!\big(\mathrm{LN}(X_{m})\big).
\end{equation}

For the spatial expert branch, inspired by Mona, we employ a multi-scale depthwise separable convolutional fusion structure that captures spatial contextual information under different receptive fields while maintaining a lightweight design. This structure performs balanced fusion of multi-scale responses along the channel dimension to capture rich spatial features, which can be formally expressed as:
\begin{equation}
\bar{X}_{m}
= \tfrac{1}{3}\!\left(
\mathrm{DW}_{3}(\tilde{X}_{m})
+ \mathrm{DW}_{5}(\tilde{X}_{m})
+ \mathrm{DW}_{7}(\tilde{X}_{m})
\right),
\end{equation}
where $\mathrm{DW}_k$ denotes a depthwise separable convolution with a kernel size of $k \times k$. The fused responses are then added to the identity mapping of the input feature, followed by an additional $1 \times 1$ convolution to achieve channel-wise linear mixing. This design enhances cross-channel feature interaction without disrupting the original representation distribution and finally projects the refined feature back to the original channel dimension:
\begin{equation}
\Delta_{m}^{\text{spatial}}
= \mathrm{Conv}_{1\times1}\!\left(
\bar{X}_{m} + \tilde{X}_{m} + \mathrm{Conv}_{1\times1}(\bar{X}_{m})
\right).
\end{equation}

The spatial-domain expert primarily relies on explicit local convolutional neighborhood modeling to capture contextual information but often struggles to effectively characterize the correlations and discrepancies among multi-modal features. As a complement, the frequency-domain expert branch explicitly captures inter-modal differences and complementarity in the spectral space. Specifically, we first apply a two-dimensional Fast Fourier Transform (FFT) to the input feature $X_m$ and divide the resulting spectral representation into low- and high-frequency components based on a predefined cutoff ratio $\rho \in (0,1)$. The computation can be formulated as:
\begin{equation}
\left\{
\begin{aligned}
\mathcal{X}_{m}^{\mathrm{low}} &= \mathcal{F}(\tilde{X}_{m}) \odot \mathcal{M}, \\[4pt]
\mathcal{X}_{m}^{\mathrm{high}} &= \mathcal{F}(\tilde{X}_{m}) \odot \big(1 - \mathcal{M}\big),
\end{aligned}
\right.
\end{equation}
where $\mathcal{F}(\cdot)$ denotes the Fourier transform operator, $\odot$ represents element-wise multiplication, and $M(\rho)$ is a low-frequency mask defined according to the cutoff ratio as follows:
\begin{equation}
\mathcal{M}(u, v) =
\begin{cases}
1, & \text{if } \max\!\left(|u - \tfrac{H}{2}|,\, |v - \tfrac{W}{2}|\right) \le \rho \tfrac{H}{2}, \\[6pt]
0, & \text{otherwise},
\end{cases}
\end{equation}
where $(u,v)$ denote the frequency coordinates. Through this masking operation, the model adaptively divides the spectral features into two complementary subspaces: the low-frequency component $X_m^{\text{low}}$ primarily represents global structures and smoothly varying illumination, whereas the high-frequency component $X_m^{\text{high}}$ emphasizes textures, edges.

Subsequently, the inverse Fast Fourier Transform (iFFT), denoted as $\mathcal{F}^{-1}$, is applied to the low- and high-frequency components to reconstruct them back into the spatial domain. The restored features are processed by a depthwise convolution and a channel attention module for lightweight encoding and adaptive reweighting, enabling the model to dynamically balance different frequency components under significant inter-modal discrepancies.
\begin{equation}
\begin{cases}
X_m^{\mathrm{low}}  = \mathrm{DW}_{3}\!\left[\mathcal{F}^{-1}\!\big(\mathcal{X}_m^{\mathrm{low}}\big)\right],\\[4pt]
X_m^{\mathrm{high}} = \mathrm{DW}_{3}\!\left[\mathcal{F}^{-1}\!\big(\mathcal{X}_m^{\mathrm{high}}\big)\right],
\end{cases}
\end{equation}
\begin{equation}
\begin{cases}
\Delta_m^{\mathrm{low}}  = \sigma\!\left(\mathrm{Conv}_{1\times1}\!\big(\mathrm{GAP}(X_m^{\mathrm{low}})\big)\right)  \odot X_m^{\mathrm{low}},\\[4pt]
\Delta_m^{\mathrm{high}} = \sigma\!\left(\mathrm{Conv}_{1\times1}\!\big(\mathrm{GAP}(X_m^{\mathrm{high}})\big) \right)  \odot X_m^{\mathrm{high}},
\end{cases}
\end{equation}

\subsubsection{Mixture of Adapters with Router} The aforementioned co-Experts model modality-specific features in both the spatial and frequency domains, while a router is introduced to dynamically learn weighting coefficients from the original visual features, adaptively regulating the combination of expert outputs to achieve coordinated feature fusion and optimal reconstruction.

Specifically, given the three feature maps processed by the spatial and frequency expert branches $\{\Delta_{m}^{\text{spatial}}, \Delta_{m}^{\text{low}}, \Delta_{m}^{\text{high}}\}$, we first construct a router generator with a  convolution to predict the importance weights of the three experts at the pixel level:

\begin{equation}
\mathbf{W}_m = \mathrm{Softmax}\!\left(\mathrm{Conv}_{1\times1}\!\left(X_m\right)\right),
\end{equation}
where the three channels of $\mathbf{W}_m$ correspond to the response contributions of the spatial-, low-, and high-frequency experts. The Softmax operation ensures that the three weights at each pixel position sum to one, thereby achieving explicit energy normalization. The fused feature is then formulated as a pixel-wise weighted summation as follows:
\begin{equation}
\Delta_m = 
\mathbf{W}_m^{(1)} \odot \Delta_m^{\mathrm{spatial}} \;+\;
\mathbf{W}_m^{(2)} \odot \Delta_m^{\mathrm{low}} \;+\;
\mathbf{W}_m^{(3)} \odot \Delta_m^{\mathrm{high}},
\end{equation}
Finally, the output of our LFM-Adapter is obtained through a residual aggregation path to retain the fundamental input representation and prevent modality over-alignment:
\begin{equation}
F_m = X_m + \Delta_m.
\end{equation}

\begin{table*}[htbp]
  \centering
  \setlength{\tabcolsep}{5pt}
  \renewcommand{\arraystretch}{1.15}
  \caption{Dataset Overview for Multimodal Object Detection Scenarios.}
  \label{tab:dataset_summary}
  \begin{adjustbox}{max width=\textwidth}
  \begin{tabular}{l c c c c c c}
    \toprule
    \textbf{Dataset} & \textbf{Image Type} & \textbf{Image Size} & \textbf{Train Set} & \textbf{Test Set} & \textbf{Classes} & \textbf{Category Names} \\
    \midrule
    VEDAI~\cite{RAZAKARIVONY2016187} & RGB + Infrared & $1024 \times 1024$ & 1,089 & 121 & 8 & car, pickup, camping car, truck, other, tractor, boat, van \\
    M3FD~\cite{liu2022target} & RGB + Thermal & $640 \times 640$ & 2,905 & 1,295 & 6 & person, car, bus, motorcycle, lamp, truck \\
    FLIR~\cite{zhang2020multispectral} & RGB + Thermal & $512 \times 640$ & 4,129 & 1,013 & 3 & bicycle, car, person \\
    DroneVehicle~\cite{9759286} & RGB + Infrared & $712 \times 840$ & 17,990 & 8,980 & 5 & car, truck, bus, van, freight car \\
    \bottomrule
  \end{tabular}
  \end{adjustbox}
\end{table*}

\section{EXPERIMENTS}

\subsection{Datasets and Evaluation Metric}

In our experiments, we employ four publicly available multimodal datasets to comprehensively evaluate the effectiveness and generalization capability of MM-DETR. The dataset configurations strictly follow the standardized benchmarks established in \cite{zhang2024rethinking, zhang2023superyolo, liu2025dual}, thereby ensuring the comparability and fairness of our experimental results. The main characteristics of each dataset are summarized in Table~\ref{tab:dataset_summary}, and detailed descriptions are provided as follows.

\subsubsection{VEDAI~\cite{RAZAKARIVONY2016187}} This dataset consists of paired RGB and infrared aerial images, containing a total of 1,246 image pairs. 
Following the established benchmark protocol, we adopt the $1024 \times 1024$-resolution version, 
where 1,089 pairs are used for training and 121 pairs are reserved for testing. 
It contains eight object categories, mainly covering ground vehicles and building-related targets.

\subsubsection{M3FD~\cite{liu2022target}} Released by Liu et al., the M3FD dataset consists of paired thermal–visible multimodal images covering six object categories.
It is designed to support object detection research under diverse scenes and illumination conditions. 
Following the experimental setup of \cite{zhang2024rethinking}, 
we split the data into 2,905 training pairs and 1,295 testing pairs, 
and uniformly resize all images to $640 \times 640$ pixels.

\subsubsection{FLIR~\cite{zhang2020multispectral}} This dataset targets traffic-scene perception tasks and contains synchronized RGB and thermal image pairs, covering three major categories. According to the official split, 4,129 image pairs are used for training and 1,013 pairs for testing, with all images resized to 640 × 640 pixels for input.

\subsubsection{DroneVehicle~\cite{9759286}} Published by Tianjin University, this multimodal UAV dataset consists of co-registered optical and infrared image pairs captured by a dual-camera system. 
It encompasses a diverse range of urban and suburban scenes under varying illumination conditions from daytime to nighttime. 
The dataset includes five annotated categories, with 17,990 image pairs for training and 8,980 for testing. 
All images are uniformly resized to $640 \times 640$ pixels before being fed into the model during experiments.

We adopt the mean Average Precision (mAP) at an Intersection over Union (IoU) threshold of 0.5 as our evaluation metric, which is widely used in various multimodal object detection tasks.

\subsection{Implementation Details}

We adopt RT-DETR\cite{zhao2024detrs} as our baseline detector and compare it with both CNN-based and Transformer-based multimodal object detection methods. 
To ensure a fair comparison, all models use ResNet-50 (pretrained on ImageNet) as the backbone whenever possible. 
In all experiments, our model is trained for up to 72 epochs, following the standard training protocol of RT-DETR. 
We employ the AdamW optimizer with a base learning rate of 0.0001, $\beta_1 = 0.9$, and $\beta_2 = 0.999$. 
All experiments are conducted on a single NVIDIA A6000 GPU with 48~GB of memory.

\subsection{Comparing with State-of-the-Arts Approaches}

To comprehensively evaluate the effectiveness of MM-DETR, we performed a comparison with various state-of-the-art multimodal object detection methods across four representative datasets. These datasets cover different application scenarios, including remote sensing imagery (VEDAI and DroneVehicle) and traffic perception scenarios (FLIR and M3FD). These datasets represent a variety of environmental conditions and viewpoints, allowing us to thoroughly assess the robustness and generalization capability of MM-DETR across different scenarios.

\begin{table*}[htbp]
  \centering
  \scriptsize
  \setlength{\tabcolsep}{4pt}
  \renewcommand{\arraystretch}{1.12}
  \begin{threeparttable}
  \caption{Experimental results (\%) on the VEDAI dataset under the RGB-IR fusion setting.}
  \label{tab:vedai_results}
  \begin{tabular}{c|c|l|cccccccc|c}
    \toprule
    Method & Detector & \multicolumn{1}{c|}{Backbone} 
    & Car & Pickup & Camping & Truck & Other & Tractor & Boat & Van & \textbf{mAP$_{50}$} \\
    \midrule
    RetinaNet~\textcolor{gray}{\footnotesize [TPAMI'20]}~\cite{lin2017focal} & RetinaNet & ResNet-50\textsuperscript{\dag} 
        & 93.36 & 87.86 & \textbf{90.61} & 83.39 & 51.91 & 72.94 & 74.22 & 68.24 & 77.82 \\
    YOLOrs~\textcolor{gray}{\footnotesize [JSTARS'21]}~\cite{sharma2020yolors} & YOLOv3 & Darknet-53\textsuperscript{\dag} 
        & 84.15 & 78.27 & 68.81 & 52.60 & 46.75 & 67.88 & 21.47 & 57.91 & 59.73 \\
    FCOS~\textcolor{gray}{\footnotesize [TPAMI'22]}~\cite{tian2019fcos} & FCOS & ResNet-50\textsuperscript{\dag} 
        & 85.86 & 79.22 & 86.08 & 90.93 & 57.27 & 75.36 & 70.53 & 80.20 & 78.20 \\
    YOLOFusion~\textcolor{gray}{\footnotesize [PR'22]}~\cite{qingyun2022cross} & YOLOv5 & CSPD-53\textsuperscript{\dag} 
        & 91.70 & 85.90 & 78.90 & 78.10 & 54.70 & 71.90 & 71.70 & 75.20 & 75.90 \\
    SuperYOLO~\textcolor{gray}{\footnotesize [TGRS'23]}~\cite{zhang2023superyolo} & YOLOv5 & CSPD-53 
        & 91.13 & 85.66 & 79.30 & 70.18 & 57.33 & 80.41 & 60.24 & 76.50 & 75.09 \\
    FFCA~\textcolor{gray}{\footnotesize [TGRS'24]}~\cite{10423050} & YOLOv5 & CSPD-53 
        & 89.60 & 85.70 & 78.70 & 75.70 & 48.60 & 81.80 & 61.50 & 67.00 & 74.80 \\
$\mathrm{C}^{2}\mathrm{Former}$~\textcolor{gray}{\footnotesize [TGRS'24]}~\cite{yuan2024c2former} & Cascade R-CNN & ResNet-50\textsuperscript{\dag} 
        & 87.20 & 80.70 & 77.60 & 79.40 & 58.40 & 72.90 & 71.40 & 75.20 & 75.70 \\
    YOLOFIV~\textcolor{gray}{\footnotesize [JSTARS'24]}~\cite{wang2024yolofiv} & YOLOv5 & CSPD-53\textsuperscript{\dag} 
        & 93.89 & 87.42 & 82.10 & 61.04 & \textbf{82.15} & 75.47 & 79.28 & 80.16 & 80.16 \\
    CAFF-DINO~\textcolor{gray}{\footnotesize [CVPR'24]}~\cite{helvig2024caff} & DINO & ResNet-50\textsuperscript{\dag} 
        & 88.15 & 83.16 & 81.14 & 69.97 & 60.64 & 71.67 & 69.59 & 68.91 & 74.15 \\
    GM-DETR~\textcolor{gray}{\footnotesize [CVPR'24]}~\cite{xiao2024gm} & RT-DETR & ResNet-50 
        & 91.15 & 87.49 & 83.43 & 89.83 & 70.92 & 81.02 & 91.87 & \textbf{94.06} & 86.22 \\
    Multi-spectral DETR~\textcolor{gray}{\footnotesize [JSTARS'25]}~\cite{10770223} & Deformable-DETR & Swin\textsuperscript{\dag} 
        & -- & -- & -- & -- & -- & -- & -- & -- & 82.20 \\
    ADMPF~\textcolor{gray}{\footnotesize [TGRS'25]}~\cite{liu2025aerial} & YOLOv8 & C2f-CSP 
        & \textbf{94.45} & \textbf{90.25} & 84.74 & 74.41 & 57.16 & 74.74 & 85.46 & 91.86 & 81.63 \\
    DPAL-P~\textcolor{gray}{\footnotesize [TGRS'25]}~\cite{liu2025dual} & FCOS & ResNet-50\textsuperscript{\dag} 
        & 88.10 & 83.88 & 86.79 & \textbf{95.53} & 67.98 & 71.07 & 87.06 & 87.05 & 83.43 
        \\
    MS2Fusion~\textcolor{gray}{\footnotesize [Inf. Fusion'26]}~\cite{SHEN2026103895} & CO-DETR & ResNet-50\textsuperscript{\dag} 
    & 93.30 & 88.50 & 80.60 & 78.90 & -- & 85.80 & 65.50 & 83.60 & 84.20 
    \\
    \midrule
    \textbf{MM-DETR (Ours)} & RT-DETR & ResNet-50 
        & 93.03 & 88.81 & 85.95 & 82.74 & 71.96 & \textbf{89.16} & \textbf{92.00} & 92.34 & \textbf{87.06} \\
    \bottomrule
  \end{tabular}
  \begin{tablenotes}[flushleft]
    \footnotesize
    \item[{\dag}] Indicates that a dual-stream backbone is employed to extract modality-specific features.
  \end{tablenotes}
  \end{threeparttable}
\end{table*}

\begin{table}[htbp]
  \centering
  \footnotesize
  \caption{Experimental results (\%) on the DroneVehicle dataset under the RGB-IR fusion setting.}
  \label{tab:dv_results}
  \resizebox{\columnwidth}{!}{%
  \begin{tabular}{c|ccccc|c}
    \toprule
    Method & Car & Truck & Freight & Bus & Van & \textbf{mAP$_{50}$} \\
    \midrule
    % ----------------- 2019 -----------------
    FCOS~\cite{tian2019fcos}
        & 94.32 & 70.30 & 53.18 & 92.26 & 42.72 & 70.55 \\
    % ----------------- 2022 -----------------
    CGRP~\cite{wang2022unsupervised}
        & 89.90 & 66.40 & 60.80 & 88.90 & 51.30 & 71.40 \\
    % ----------------- 2024 -----------------
    ICA-FCOS~\cite{shen2024icafusion}
        & 81.60 & 56.00 & 33.30 & 85.70 & 31.80 & 57.70 \\
    $\mathrm{C}^{2}\mathrm{Former}$~\cite{yuan2024c2former}
        & 84.10 & 69.60 & 60.60 & 89.50 & 57.90 & 71.60 \\
    LFMDet~\cite{sun2024low}
        & 82.20 & 73.60 & 59.60 & 86.60 & 57.00 & 71.50 \\
    CAFF-DINO~\cite{helvig2024caff}
        & 94.05 & 82.99 & 66.94 & 95.62 & 65.67 & 81.07 \\
    GM-DETR~\cite{xiao2024gm}
        & 94.00 & 82.89 & 68.69 & 95.69 & 66.94 & 81.64 \\
    Multi-spectral DETR~\cite{10770223}
        & -- & -- & -- & -- & -- & 76.90 \\
    % ----------------- 2025 -----------------
    DPAL-P~\cite{liu2025dual}
        & \textbf{95.25} & 74.82 & 58.69 & 94.16 & 51.60 & 74.91 \\
    WaveMamba~\cite{zhu2025wavemamba}
        & 95.00 & 80.40 & 68.50 & 90.60 & 64.50 & 79.80 \\
    % ----------------- 2026 -----------------
    SM3Det~\cite{Li2026SM3Det}
        & -- & -- & -- & -- & -- & 77.99 \\
    \midrule
    % ----------------- Ours -----------------
    \textbf{MM-DETR (Ours)}
        & 93.91 & \textbf{84.14} & \textbf{69.04} 
        & \textbf{95.98} & \textbf{68.46} 
        & \textbf{82.31} \\
    \bottomrule
  \end{tabular}
  }%
\end{table}

\subsubsection{Results on VEDAI and DroneVehicle} We first evaluated the detection performance in remote sensing scenarios. Table~\ref{tab:vedai_results} presents the validation results on the VEDAI dataset, where MM-DETR achieved an mAP$_{50}$ of 87.06\%, significantly outperforming existing methods. Notably, MM-DETR excels in challenging categories such as "Tractor," demonstrating the model's enhanced ability to effectively handle objects with complex structures and subtle visual features. Table~\ref{tab:dv_results} shows the results for the DroneVehicle dataset, where MM-DETR achieved an mAP$_{50}$ of 82.31\%, significantly surpassing CNN-based architecture methods. This highlights that DETR-based methods excel in detecting small and densely distributed vehicles across various categories. Our approach outperforms the best GM-DETR method by 0.64\% mAP while maintaining a smaller parameter size and computational footprint, further demonstrating the superior modality-specific feature extraction capability of our model, along with an effective and lightweight modality fusion mechanism.

\begin{table}[htbp]
  \centering
  \scriptsize
  \setlength{\tabcolsep}{3pt}
  \renewcommand{\arraystretch}{1.12}
  \begin{threeparttable}
  \caption{Experimental results (\%) on the M3FD dataset under the RGB-T fusion setting.}
  \label{tab:m3fd_results}
  \begin{tabular}{c|cccccc|c}
    \toprule
    Method & Person & Car & Bus & Motorcycle & Lamp & Truck & \textbf{mAP$_{50}$} \\
    \midrule
    % ----------------- 2017 -----------------
    RetinaNet~\cite{lin2017focal} 
        & 60.10 & 77.27 & 61.63 & 45.42 & 25.67 & 51.80 & 53.63 \\
    % ----------------- 2019 -----------------
    FCOS~\cite{tian2019fcos} 
        & 69.99 & 81.92 & 63.25 & 41.58 & 42.72 & 50.50 & 58.33 \\
    % ----------------- 2022 -----------------
    GFL~\cite{li2022generalized} 
        & 65.37 & 79.83 & 61.20 & 37.00 & 34.80 & 48.73 & 54.47 \\
    % ----------------- 2024 -----------------
    ShaPE~\cite{zhang2024rethinking} 
        & 65.80 & 79.10 & 62.33 & 41.33 & 30.80 & 53.67 & 55.50 \\
    EME~\cite{zhang2024rethinking} 
        & 68.43 & 81.23 & 63.37 & 43.90 & 35.77 & 53.53 & 57.70 \\
    CAFF-DINO~\cite{helvig2024caff} 
        & 82.10 & 87.66 & 51.62 & 57.13 & 60.24 & 59.67 & 66.41 \\
    GM-DETR~\cite{xiao2024gm} 
        & 80.69 & 88.09 & 74.05 & 60.73 & 64.63 & \textbf{66.11} & 72.41 \\
    % ----------------- 2025 -----------------
    DPAL-P~\cite{liu2025dual} 
        & 71.54 & 83.31 & 68.96 & 44.79 & 44.45 & 57.38 & 61.74 \\
    \midrule
    % ----------------- Ours -----------------
    \textbf{MM-DETR (Ours)} 
        & \textbf{82.53} 
        & \textbf{89.05} 
        & \textbf{74.30} 
        & \textbf{62.39} 
        & \textbf{66.35} 
        & 65.63 
        & \textbf{73.39} \\
    \bottomrule
  \end{tabular}
  \end{threeparttable}
\end{table}

\begin{table}[htbp]
  \centering
  \footnotesize
  \setlength{\tabcolsep}{4pt}
  \renewcommand{\arraystretch}{1.12}
  \begin{threeparttable}
  \caption{Experimental results (\%) on the FLIR dataset under the RGB-T fusion setting.}
  \label{tab:flir_results}
  \begin{tabular}{c|ccc|c}
    \toprule
    Method & Bicycle & Car & Person & \textbf{mAP$_{50}$} \\
    \midrule
    % ----------------- 2017 -----------------
    RetinaNet~\cite{lin2017focal}
      & 67.60 & 85.17 & 61.93 & 71.57 \\
    % ----------------- 2019 -----------------
    FCOS~\cite{tian2019fcos}
      & 56.89 & 83.47 & 76.73 & 72.37 \\
    % ----------------- 2022 -----------------
    CMPD~\cite{li2022confidence}
      & 59.87 & 78.11 & 69.64 & 69.35 \\
    % ----------------- 2024 -----------------
    ICA-FCOS~\cite{shen2024icafusion}
      & -- & -- & -- & 71.70 \\
    EME~\cite{zhang2024rethinking}
      & 69.23 & 85.10 & 62.27 & 72.23 \\
    CAFF-DINO~\cite{helvig2024caff}
      & 66.18 & 89.94 & 84.15 & 80.09 \\
    GM-DETR~\cite{xiao2024gm}
      & 70.46 & 90.96 & 85.81 & 82.41 \\
    % ----------------- 2025 -----------------
    DPAL-P~\cite{liu2025dual}
      & 63.14 & 85.72 & 78.99 & 75.95 \\
    \midrule
    % ----------------- Ours -----------------
    \textbf{MM-DETR (Ours)}
      & \textbf{73.13} & \textbf{91.26} & \textbf{86.37} & \textbf{83.59} \\
    \bottomrule
  \end{tabular}
  \end{threeparttable}
\end{table}

\subsubsection{Results on M3FD and FLIR} We further evaluated MM-DETR's performance in traffic perception scenarios to validate its effectiveness and robustness. Table~\ref{tab:m3fd_results} presents the validation results on the M3FD dataset, where MM-DETR achieved an mAP${50}$ of 73.39\%, surpassing the best method by 0.98 percentage points. In detail, MM-DETR performed exceptionally well in categories such as "Person" and "Lamp," demonstrating the model's ability to effectively handle diverse and multi-scale objects in traffic scenarios. As shown in Table~\ref{tab:flir_results}, on the FLIR dataset, MM-DETR achieved an mAP${50}$ of 83.59\%, significantly outperforming other methods and achieving excellent results across all categories. The results from both datasets highlight that our proposed method maintains outstanding performance in natural traffic environments. Finally, our proposed method strikes a favorable balance between performance and model parameters while maintaining a low computational cost, demonstrating its strong potential for real-world deployment.

\subsection{Ablation Studies }

The ablation experiments are conducted by replacing or removing individual components to comprehensively evaluate the contribution of each module in the proposed MM-DETR. Then, we focus on analyzing the proposed MDF-Encoder, exploring the trade-off between its performance gain and the number of parameters of the modality-completion branch. Finally, we investigate the effectiveness of the LFM-Adapter, as well as the impact of its hyperparameters, including the cutoff frequency and projector dimension, on detection performance. All experiments are conducted across multiple scenarios to demonstrate the effectiveness and generalization capability of the proposed method.

\subsubsection{Effectiveness of each component} Table~\ref{tab:ablation_multimodal} summarizes the contribution of each component. We first observe that the infrared modality alone consistently outperforms the RGB-only counterpart on both datasets, indicating that thermal imaging provides more reliable visual cues under low-light or complex environmental conditions. Building on this, simply concatenating RGB and IR features already leads to a substantial performance boost, as shown in the third row of Table, which demonstrates the inherent complementarity between the two modalities.

With the CEI module incorporated, modality-shared semantics are further enhanced while redundant information is effectively suppressed, leading to consistent performance improvements. Incorporating the MPF module additionally introduces a lightweight, fine-grained residual completion mechanism centered on modality compensation, enabling the detector to achieve the best performance of 87.29\% and 73.53\% mAP$_{50}$ on VEDAI and M3FD, respectively, corresponding to improvements of +2.55\% and +1.41\% over naive concatenation.

Finally, the LFM-Adapter replaces the dual-stream backbone with a shared one and leverages a spatial–frequency co-expert module together with a pixel-wise router to extract modality-specific features more efficiently, thereby reducing the overall parameter count from 71.12M to 50.01M. Despite this substantial simplification, only a negligible performance drop of approximately 0.2\% mAP is observed, highlighting the excellent parameter efficiency and strong deployment potential of the proposed design.

\begin{table}[t]
  \centering
  \setlength{\tabcolsep}{6pt} % 调整列间距
  \renewcommand{\arraystretch}{1.15} % 调整行间距
  \caption{Ablation study on each component of the proposed MM-DETR.}
  \label{tab:ablation_multimodal}
  \begin{adjustbox}{max width=\columnwidth}
  \begin{tabular}{c c c c | c c c}
    \toprule
    \multirow{2}{*}{\textbf{Modality}} & 
    \multicolumn{2}{c}{\textbf{MDF-Encoder}} & 
    \multirow{2}{*}{\textbf{LFM-Adapter}} & 
    \multirow{2}{*}{\textbf{Params (M)}} &
    \multicolumn{2}{c}{\textbf{mAP$_{50}$}} \\
    \cmidrule(lr){2-3}\cmidrule(lr){6-7}
     & CEI & MPF & & & VEDAI & M3FD \\
    \midrule
    I              &            &            &            & 42.74 & 82.24 & 61.60 \\
    R              &            &            &            & 42.74 & 83.17 & 67.38 \\
    \midrule
    I + R          &            &            &            & 67.13 & 84.74 & 72.12 \\
    I + R          & \ding{51}  &            &            & 67.73 & 85.73 & 72.76 \\
    I + R          & \ding{51}  & \ding{51}  &            & 71.12 & \textbf{87.29} & \textbf{73.53} \\
    \rowcolor{gray!15}
    I + R          & \ding{51}  & \ding{51}  & \ding{51}  & 50.01 & 87.06 & 73.39 \\
    \bottomrule
  \end{tabular}
  \end{adjustbox}
\end{table}

\subsubsection{Impact of modality-completion branch design\label{sec:ablation_mpf_branch}}  Table~\ref{tab:ablation_branches} analyzes the impact of different modality-completion branch configurations. Removing the MPF module results in clear performance degradation on both datasets, validating the necessity of modality completion. Enabling only the IR-side completion branch improves performance from 85.73\% to 87.29\% on VEDAI and from 72.76\% to 73.53\% on M3FD, indicating that IR cues play a more essential role in compensating modality discrepancies under remote sensing conditions. In comparison, the RGB-side completion branch yields marginal gains, suggesting that visible cues are more vulnerable to degradation in complex imaging environments. Although activating both branches achieves the highest accuracy on VEDAI, the additional complexity does not translate into consistent benefits on M3FD. This is primarily because simultaneously performing completion for both modalities introduces redundant or even conflicting information, hampering the preservation of modality-specific characteristics. Therefore, considering the balance between performance gain and model complexity, the IR-only completion branch is adopted as the default configuration in our final model.

\begin{table}[t]
  \centering
  \scriptsize
  \setlength{\tabcolsep}{6pt}
  \renewcommand{\arraystretch}{1.1}
  \caption{Ablation study on the modality-completion branches.}
  \label{tab:ablation_branches}
  \begin{adjustbox}{max width=\columnwidth}
  \begin{tabular}{c | c c c}
    \toprule
    \multirow{2}{*}{\textbf{Variant}} & \multirow{2}{*}{\textbf{Params (M)}} & \multicolumn{2}{c}{\textbf{mAP$_{50}$}} \\
    \cmidrule(lr){3-4}
     &  & VEDAI & M3FD \\
    \midrule
    w/o MPF          & 68.28 & 85.73 & 72.76 \\
    +R Branch         & 71.12 & 86.50 & 73.36 \\
    \rowcolor{gray!15}
    +I Branch         & 71.12 & 87.29 & \textbf{73.53} \\
    +Both Branches    & 73.96 & \textbf{87.34} & 73.28 \\
    \bottomrule
  \end{tabular}
  \end{adjustbox}
\end{table}

\subsubsection{Effectiveness of the frequency expert and sensitivity to cutoff frequency} Table~\ref{tab:ablation_rho} shows that the proposed LFM-Adapter achieves significant performance improvements over the Shared and Mona baselines while introducing only a negligible number of additional parameters. Compared with Mona-style spatial adapters, the proposed spatial–frequency co-expert design yields +0.34\% mAP on VEDAI and +1.14\% mAP on M3FD with almost the same parameter budget, demonstrating the benefit of explicitly modeling frequency-domain modality discrepancies. Moreover, we further investigate the influence of the cutoff frequency $\rho$. The best results are obtained when $\rho = 0.5$, achieving 87.06\% and 73.39\% mAP$_{50}$ on VEDAI and M3FD, respectively, which indicates a well-balanced modeling of low-frequency structural information and high-frequency texture details. Overall, these results demonstrate that the frequency-domain expert plays an essential role in leveraging cross-modal complementarity, and the proposed LFM-Adapter consistently yields stable improvements across datasets and different cutoff configurations, showing strong generalization ability and deployment potential.

\begin{table}[hbp]
  \centering
  \scriptsize
  \setlength{\tabcolsep}{6pt}
  \renewcommand{\arraystretch}{1.1}
  \caption{Ablation study on different $\rho$ values in the LFM-Adapter.}
  \label{tab:ablation_rho}
  \begin{adjustbox}{max width=\columnwidth}
  \begin{tabular}{l c | c c c}
    \toprule
    \multirow{2}{*}{\textbf{Method}} 
    & \multirow{2}{*}{\boldmath{$\rho$}} 
    & \multirow{2}{*}{\textbf{Params (M)}} 
    & \multicolumn{2}{c}{\textbf{mAP$_{50}$}} \\
    \cmidrule(lr){4-5}
     &  &  & VEDAI & M3FD \\
    \midrule
    Shared         & --   & 47.64 & 85.51 & 71.46 \\
    Mona~\cite{yin20255}           & --   & 49.87 & 86.72 & 72.25 \\
    \midrule
    \multirow{4}{*}{LFM-Adapter} 
    & 0.3 & \multirow{4}{*}{50.01} & 86.49 & 72.81 \\
    & 0.4 &  & 86.97 & 72.69 \\
    & \cellcolor{gray!15}\textbf{0.5} &  & \cellcolor{gray!15}\textbf{87.06} & \cellcolor{gray!15}\textbf{73.39} \\
    & 0.6 &  & 86.81 & 73.18 \\
    \bottomrule
  \end{tabular}
  \end{adjustbox}
\end{table}

\subsubsection{Effect of adapter dimension} 
\begin{figure}[htbp]
\centering
\includegraphics[width=\linewidth]{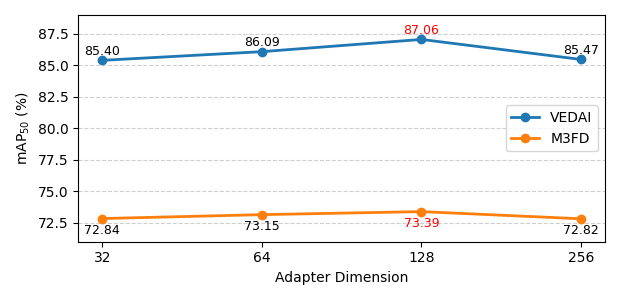}
\caption{Parameter sensitivity analysis of adapter projector dimension. Results on multiple multimodal benchmarks demonstrate that the proposed LFM-Adapter remains robust across different projector dimensions.}
\label{fig_adapter_dim}
\end{figure}

We further investigate the effect of the adapter dimension on detection performance, as illustrated in Fig.~\ref{fig_adapter_dim}. Increasing the dimension from 32 to 128 leads to consistent performance improvements on both VEDAI and M3FD, since a higher-dimensional feature space can encode richer modality-specific information and better preserve fine-grained semantic cues. However, when the dimension is increased to 256, a slight performance drop is observed on both datasets, likely due to the introduction of redundant that cause overfitting and reduce robustness. Overall, a dimension of 128 provides the best balance between representation capacity and regularization, yielding the highest performance across both multimodal benchmarks.

\subsection{Analysis and Visualization}

\subsubsection{Performance–efficiency comparison} We further present a comprehensive comparison between the proposed MM-DETR and representative CNN-based and Transformer-based multimodal detectors in terms of computational efficiency and detection performance, as shown in Table~\ref{tab:comp_Performance–efficiency}. Obviously, dual-stream backbone designs inevitably incur a substantial increase in parameter count, as two independent feature extractors must be maintained. Although GM-DETR adopts a shared backbone, it substantially increases the channel width of its encoder to strengthen representational capacity, which still leads to a relatively large model size. Benefiting from the proposed LFM-Adapter, MM-DETR achieves effective modality-specific modeling with no redundant parameters, thereby greatly improving its deployability.

Moreover, Transformer-based methods further improve multimodal detection performance due to their strong global modeling capability, yet they typically incur considerably higher FLOPs. The emerging Mamba architecture models long-range dependencies with linear complexity, offering a more favorable performance–efficiency trade-off compared with traditional attention or deformable convolutions. Benefiting from our lightweight Mamba-based fusion design, the proposed MM-DETR attains the best balance among model size, runtime efficiency, and detection performance, showing strong potential for real-world deployment.

\begin{table}[htbp]
  \centering
  \scriptsize
  \setlength{\tabcolsep}{6pt}
  \renewcommand{\arraystretch}{1.15}
  \caption{Comparison of Computational Efficiency and Detection Accuracy among Different Detectors on the DroneVehicle Dataset.}
  \label{tab:comp_Performance–efficiency}
  \begin{adjustbox}{max width=\columnwidth}
  \begin{tabular}{l c c c c c}
    \toprule
    \textbf{Method} 
    & \textbf{Params (M)} 
    & \textbf{FLOPs (G)} 
    & \textbf{FPS} 
    & \textbf{mAP$_{50}$} 
    & \textbf{Type} \\
    \midrule
    $\mathrm{C}^{2}\mathrm{Former}$~\cite{yuan2024c2former}
        & 100.8 
        & \textbf{89.9}
        & --   
        & 71.60 
        & RCNN-based \\
    DPAL-P~\cite{liu2025dual}         
        & 96.7   
        & 149.5   
        & 5.9
        & 74.91
        & FCOS-based \\
    WaveMamba~\cite{zhu2025wavemamba}         
        & 69.1    
        & --     
        & 25.0
        & 79.80
        & YOLO-based \\
    CAFF-DINO~\cite{helvig2024caff}        
        & 196.1
        & 367.7
        & 6.6
        & 81.07
        & DETR-based \\
    GM-DETR~\cite{xiao2024gm}          
        & 60.0
        & 140.0
        & 25.7
        & 81.64
        & DETR-based \\
    \midrule
    \rowcolor{gray!12}
    \textbf{MM-DETR (Ours)}    
        & \textbf{50.0} 
        & 122.2
        & \textbf{27.3}  
        & \textbf{82.31} 
        & DETR-based \\
    \bottomrule
  \end{tabular}
  \end{adjustbox}
\end{table}

\subsubsection{Analysis of training strategy variations} Typical PEFT methods insert lightweight adapters into a frozen backbone to efficiently adapt models to new tasks. Motivated by this paradigm, we investigate two training strategies for multimodal object detection: (1) a staged training scheme, where the backbone is first trained alone and the adapters are subsequently fine-tuned; (2) an end-to-end joint training scheme, where both components are optimized simultaneously. As shown in Table~\ref{tab:train_strategy_variants}, the staged training strategy yields clearly inferior performance compared with joint optimization. The primary reason is that when the backbone is optimized independently without the adapters, it tends to over-align RGB and IR features, collapsing them into an excessively unified representation space and suppressing modality-specific cues. Consequently, once the adapters are introduced in the second stage, the residual modality discrepancies become too limited to model, leading to optimization instability and degraded representations, often resulting in performance that is even lower than the shared-backbone baseline. In contrast, end-to-end joint training enables the backbone and adapters to co-adapt throughout learning, ensuring that both modality alignment and modality-specific enhancement are optimized in a unified manner. This training paradigm proves particularly effective for multimodal object detection and results in substantially improved performance.

\begin{table}[htbp]
\centering
\scriptsize
\setlength{\tabcolsep}{8pt}
\renewcommand{\arraystretch}{1.25}
\caption{Comparison of different training strategies on VEDAI and M3FD (\textit{mAP$_{50}$}).}
\label{tab:train_strategy_variants}
\begin{tabular}{l|c|c}
\toprule
\textbf{Training Strategy} 
& \textbf{VEDAI} 
& \textbf{M3FD} \\
\midrule
Shared 
& 85.51 & 71.46 \\

\midrule
Staged Training 
& 85.12 & 71.10 \\

\rowcolor{gray!15}
Joint Training 
& \textbf{87.06} & \textbf{73.39} \\
\bottomrule
\end{tabular}
\end{table}

\begin{figure}[htbp]
\centering
\includegraphics[width=9cm]{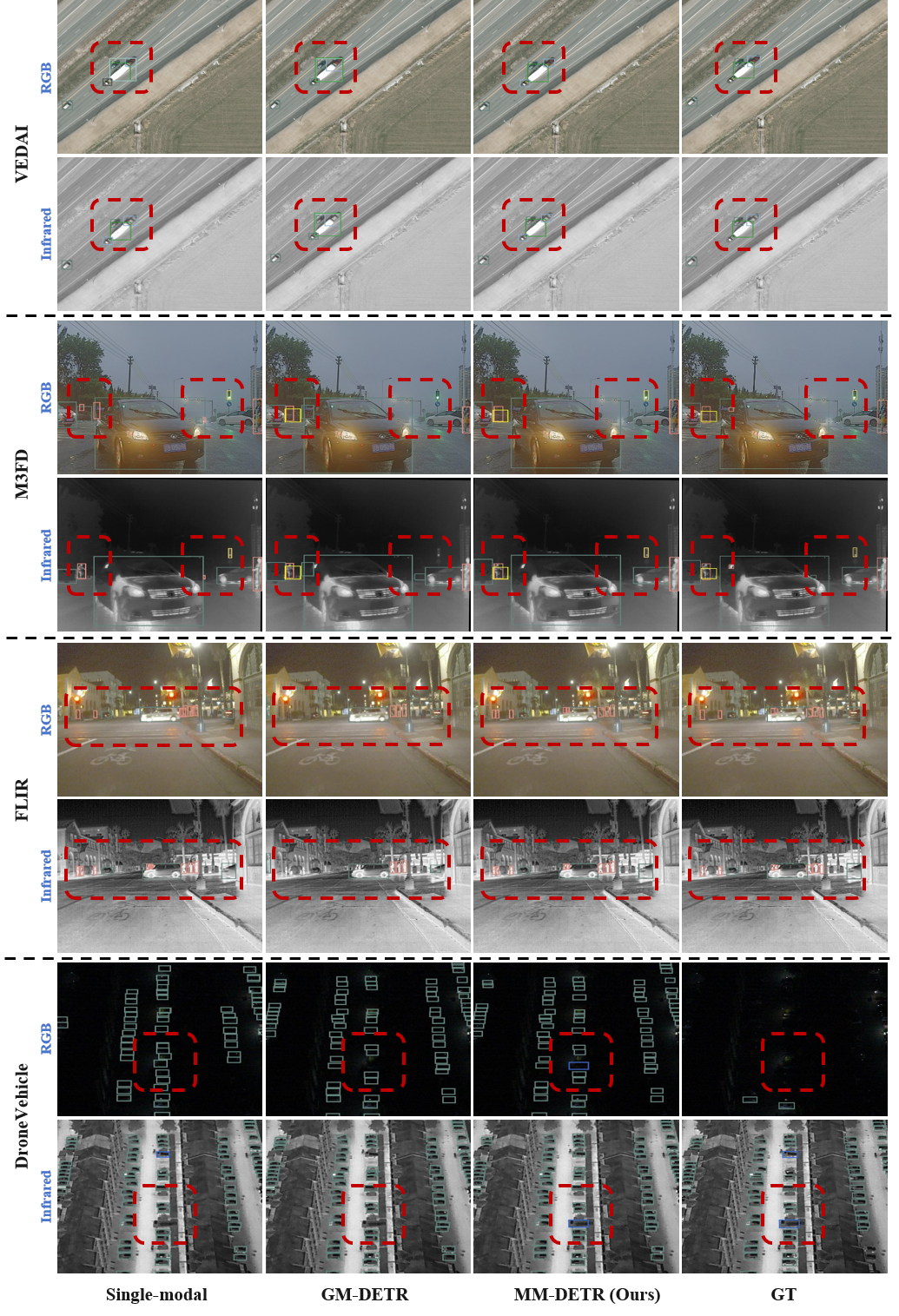}
\caption{Our detection results are compared with state-of-the-art methods. Different object categories are marked using distinct colors, and the confidence threshold for visualization is set to 0.3. “Single-modal” denotes the base detector trained using only single-modality data, whereas the other labels correspond to their respective methods. “GT” indicates the ground-truth annotations.}
\label{detect_result}
\end{figure}

\subsubsection{Detection results} We present qualitative comparisons of MM-DETR on all four benchmark datasets and compare its results against those of the single-modal baseline and the state-of-the-art multimodal method GM-DETR, as illustrated in Fig.~\ref{detect_result}. On the VEDAI dataset, MM-DETR produces more accurately localized bounding boxes, particularly in scenes containing densely distributed vehicles. This indicates that the MDF-Encoder enhances cross-modal semantic alignment, while the LFM-Adapter preserves modality-specific cues, leading to a more comprehensive RGB–IR structural fusion. For the nighttime and low-visibility scenes in the M3FD and FLIR datasets, the optical modality experiences severe degradation, whereas the infrared modality remains relatively stable under challenging illumination conditions. GM-DETR, which primarily relies on a simple concatenation-based fusion scheme, fails to sufficiently leverage infrared cues, resulting in missed detections of small and densely distributed objects. In contrast, MM-DETR employs an IR-driven completion branch together with a region-aware SS2D module to inject key structural residuals from the infrared stream into the fusion pathway along the bidirectional pyramid. This design facilitates a more effective fusion of modality-specific representations by fully exploiting the informative cues in the infrared modality, thereby substantially improving nighttime small-object detection. On the DroneVehicle dataset, MM-DETR retrieves more true positives in crowded parking lots and complex urban scenes while reducing bounding-box drift and duplicate predictions. These improvements confirm the model’s strong capability for small-object modeling and fine-grained multimodal fusion in dense remote sensing scenarios. The qualitative observations align well with the quantitative results, demonstrating the effectiveness and robustness of the proposed MM-DETR in multimodal object detection tasks.

\section{Conclusion}

In this work, we propose MM-DETR, an efficient multimodal detection transformer designed for robust multimodal object detection. To address the performance–efficiency trade-offs inherent in attention- and deformable-convolution-based fusion methods, we develop the Mamba-based Dual-granularity Fusion Encoder (MDF-Encoder). This encoder reformulates global interaction as a channel-wise dynamic gating process, enabling efficient cross-modal global modeling with linear complexity. In addition, we reinterpret multimodal fusion as a modality-completion problem and introduce a modality-completion branch based on region-aware 2D selective scanning along the bidirectional pyramid pathway, achieving lightweight yet highly effective region-level multimodal fusion. Moreover, we present the Lightweight Frequency-aware Modality Adapter (LFM-Adapter), which employs a spatial–frequency co-expert structure to more precisely capture modality-specific characteristics, while a pixel-wise router dynamically modulates expert contributions. This design enables efficient modality-specific representation learning without relying on a dual-stream backbone, substantially reducing parameter redundancy. Extensive experiments across multiple multimodal detection benchmarks demonstrate the superior performance of MM-DETR, confirming the effectiveness and robustness of the proposed framework.

%\begin{thebibliography}{1}
\bibliographystyle{IEEEtran}
\bibliography{References.bib}
%\end{thebibliography}

\vfill

\end{document}